\documentclass[10pt,twoside,twocolumn,english,conference]{IEEEtran}
\usepackage{mathptmx}

\usepackage[T1]{fontenc}
\usepackage[latin9]{inputenc}
\usepackage[letterpaper]{geometry}
\usepackage{babel}
\usepackage{graphicx}
\usepackage[unicode=true,
 bookmarks=true,bookmarksnumbered=false,bookmarksopen=false,
 breaklinks=false,pdfborder={0 0 1},backref=false,colorlinks=false]
 {hyperref}
\hypersetup{pdftitle={Real-Time RGBD Odometry for Fused-State Navigation Systems},
 pdfauthor={Andrew R. Willis and Kevin M. Brink},
 allcolors=blue}

\makeatletter

\providecommand{\tabularnewline}{\\}

\@ifundefined{showcaptionsetup}{}{%
 \PassOptionsToPackage{caption=false}{subfig}}
\usepackage{subfig}
\makeatother

\begin{document}
\author{ 
\IEEEauthorblockN{Andrew R. Willis} 
\IEEEauthorblockA{
	Department of Electrical and Computer Engineering\\
	University of North Carolina at Charlotte\\ 
	Charlotte, NC 28223-0001\\
	Email: arwillis@uncc.edu
}

\and

\IEEEauthorblockN{Kevin M. Brink} 
\IEEEauthorblockA{
	Air Force Research Laboratory\\
	Munitions Directorate \\
    Eglin AFB, FL, 32542 \\
	Email: kevin.brink@us.af.mil
}}

\title{Real-Time RGBD Odometry for Fused-State Navigation Systems}
\maketitle
\begin{abstract}
This article describes an algorithm that provides visual odometry
estimates from sequential pairs of RGBD images. The key contribution
of this article on RGBD odometry is that it provides both an odometry
estimate and a covariance for the odometry parameters in real-time
via a representative covariance matrix. Accurate, real-time parameter
covariance is essential to effectively fuse odometry measurements
into most navigation systems. To date, this topic has seen little
treatment in research which limits the impact existing RGBD odometry
approaches have for localization in these systems. Covariance estimates
are obtained via a statistical perturbation approach motivated by
real-world models of RGBD sensor measurement noise. Results discuss
the accuracy of our RGBD odometry approach with respect to ground
truth obtained from a motion capture system and characterizes the
suitability of this approach for estimating the true RGBD odometry
parameter uncertainty.
\end{abstract}


\section{Introduction}

Visual image sensors, e.g., digital cameras, have become an important
component to many autonomous and semi-autonomous navigational systems.
In many of these contexts, image processing algorithms process sensed
images to generate autonomous odometry estimates for the vehicle;
referred to as visual odometry. Early work on visual odometry dates
back more than 20 years and applications of this approach are pervasive
in commercial, governmental and military navigation systems. Algorithms
for visual odometry compute the change in pose, i.e., position and
orientation, of the robot as a function of observed changes in sequential
images. As such, visual odometry estimates allow vehicles to autonomously
track their position for mapping, localization or both of these tasks
simultaneously, i.e., the Simultaneous Localization and Mapping (SLAM)
applications. The small form factor, low power consumption and powerful
imaging capabilities of imaging sensors make them an attractive as
a source of odometry in GPS-denied contexts and for ground-based,
underground, indoor and near ground navigation. 

While visual odometry using conventional digital cameras is a classical
topic, recent introduction of commercial sensors that capture both
color (RGB) images and depth (D) simultaneously, referred to as RGBD
sensors, has created significant interest for their potential application
for autonomous 3D mapping and odometry. These sensors capture color-attributed
$(X,Y,Z)$ surface data at ranges up to 6m. from a 58.5H x 46.6V degree
field of view. The angular resolution of the sensor is \textasciitilde{}5
pixels per degree and sensed images are generated at rates up to 30
Hz.

As in other visual odometry approaches, RGBD-based odometry methods
solve for the motion of the camera, referred to as ego-motion, using
data from a time-sequence of sensed images. When the geometric relationship
between the camera and vehicle body frame is known, estimated camera
motions also serve as estimates for vehicle odometry.

Algorithms that compute odometry from RGBD images typically match
corresponding 3D $(X,Y,Z)$ surface measurements obtained from sequential
frames of sensed RGBD image data. Since the image data is measured
with respect to the camera coordinate system, camera motions induce
changes in the $(X,Y,Z)$ surface coordinates of stationary objects
in the camera view. The transformation that aligns corresponding $(X,Y,Z)$
measurements in two frames indicates how the camera orientation and
position changed between these frames which, as mentioned previously,
also specifies vehicle odometry.

\begin{figure}
\noindent \begin{centering}
\includegraphics[height=2.3in]{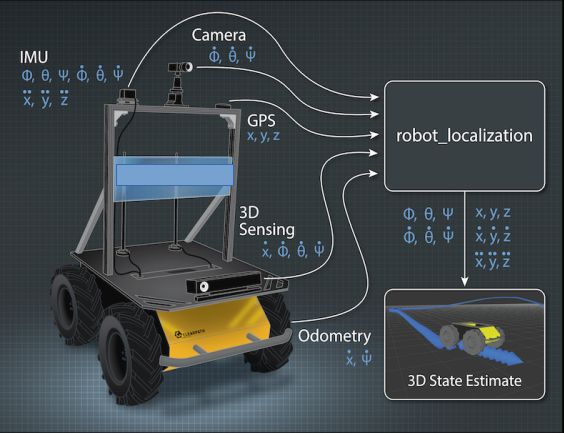}
\par\end{centering}
\caption{\label{fig:Processing-pipeline}A typical fused-state estimator described
in \cite{MooreStouchKeneralizedEkf2014} is shown. Fused-state navigation
systems integrate visual odometry, e.g., RGBD odometry, with odometry
from sensors such as IMUs, GPS, and wheel odometry to provide a single
3D state estimate for the vehicle. Accurate fused-state estimation
relies heavily on knowledge of uncertainty in each sensor measurement
to determine the best possible state estimate. This article introduces
an RGBD odometry algorithm that provides both odometry values and
uncertainties for the estimated odometry values which are critical
to this important application.}
\end{figure}

While several methods for RGBD odometry estimation exist in the literature,
published work on this topic lacks sensitivity to the larger issue
of integrating RGBD odometry estimates into autonomous navigation
systems. Researchers and practitioners acknowledge that visual odometry
can provide highly inaccurate odometry estimates; especially when
incorrect correspondences are computed. As shown in Fig. \ref{fig:Processing-pipeline},
navigation systems cope with this issue by adding odometry sensors
in the form of wheel-based odometry (ground vehicles) or Inertial
Measurement Units (IMUs) (aerial vehicles) and subsequently fusing
this data into a 3D state estimate for the vehicle. Fused state estimates
are typically produced by filtering sensor data using an Extended
Kalman Filter (EKF) \cite{MooreStouchKeneralizedEkf2014} or particle
filter. Regardless of the filter type, fused-state estimation requires
uncertainty estimates for each sensor measurement that, for an EKF,
re-weight the measurements by attributing values having low uncertainty
larger weight than other values. As such, navigational systems that
seek to leverage RGBD odometry \emph{require }uncertainty estimates
for these parameters.

For EKF-based navigation systems, absence of the RGBD odometry parameter
covariance implies that these systems include hand-specified or ad-hoc
estimates for the covariance of these parameters. For example, one
potential approach in this context is to assume a worst-case covariance
for the translational and orientation parameters via a user-specified
constant-valued diagonal covariance matrix. Yet, without means for
outlier rejection, artificially large covariances must be assumed
to cope with the possibility of potentially inaccurate RGBD odometry
parameters. When outlier rejection is applied, ad-hoc methods must
be used that dynamically change the parameter covariances which artificially
adjusts the weight of RGBD odometry estimates. While such ad-hoc approaches
may be more effective in leveraging RGBD odometry values, the artificially
chosen covariances do not characterize the true parameter uncertainty.
As a result, even sophisticated ad-hoc methods will significantly
limit the beneficial impact that RGBD odometry can have for system
navigation.

\emph{The key contribution of this article on RGBD odometry is that
it provides both an odometry estimate and a covariance for the odometry
parameters in real-time via a representative covariance matrix.} This
article proposes a statistical perturbation-based approach based on
observed RGBD measurement noise to compute the parameter covariance
estimates essential for efficiently leveraging estimated RGBD odometry
values in navigation systems. Results discuss the accuracy of our
RGBD odometry approach with respect to ground truth obtained from
a motion capture system and characterizes the suitability of this
approach for estimating the true RGBD odometry parameter uncertainty.

This article is divided into six sections as follows: \S~\ref{sec:Prior-Work}
discusses prior work on RGBD odometry related to this article, \S~\ref{sec:Background-Information}
discusses background information critical to the explanation of the
odometry algorithm, \S~\ref{sec:Methodology} discusses the RGBD
odometry and covariance parameter estimation algorithm, \S~\ref{sec:Results}
presents results for experimental odometry parameter and covariance
estimation and performance of the odometry estimation algorithm versus
ground truth data obtained via motion capture, \S~\ref{sec:Conclusion}
summarizes the impact of this work and discusses aspects of this work
that motivate future research. 

\section{\label{sec:Prior-Work}Prior Work}

Since the commercial introduction of RGBD sensors, researchers have
been developing methods to leverage the additional depth data to improve
state-of-the-art for navigation and mapping. We touch on the most
popular approaches for RGBD odometry here which 

In \cite{6225151} authors compute odometry from RGBD images by expediting
the ICP algorithm \cite{121791}. Their approach computes and matches
edges from the RGB data in sequential frames to quickly find matching
3D points from the depth image. The depth images are then aligned
by aligning 3D points at edge locations and uses the resulting transform
as an initial guess to accelerate the convergence of the computationally-costly
generalized ICP algorithm, \cite{Segal-RSS-09}, which solves for
the final estimate using a point-to-plane error metric.

In \cite{7139931} a MAV is navigated in dark environments using 6DoF
localization provided by fusing a fast odometry algorithm and generalizing
a ground-based Monte Carlo Localization approach \cite{Thrun:2005:PR:1121596}
for their 6DoF MAV application. Real-time odometry is obtained by
expressing the relative pose estimation problem as a differential
equation on each 3D point that models the change in the 3D point location
as a function of the changing pose parameters. The approach is made
real-time by sub-sampling the image data which reduces the full-frame
(640x480) image to a much smaller (80x60) image for processing. 

In \cite{6696650} a warping function is used to map pixels in sequential
RGBD images. The warping function combines an appearance model with
a depth/geometric model to solve for the transformation parameters
that minimize the differences between measured RGB colors and depths
in the warped/aligned RGBD images.

While each of the approaches above for RGBD odometry have different
strengths and weaknesses, they all suffer from a lack of consideration
of the parameter uncertainty. As mentioned in the introduction, no
visual odometry approach has been demonstrated to be completely free
of incorrect odometry estimates.\emph{ As such, regardless of the
sophistication and accuracy of the approach, it is critical to provide
uncertainty measures for RGBD odometry estimates.} Absence of this
information severely limits the utility of this information for its
primary purpose: as a component to a fused-state navigational system.

\section{\label{sec:Background-Information}Background Information}

The proposed approach for odometry draws heavily from published results
on RGBD sensor characterization, computer vision, image processing
and 3D surface matching literature. This section includes aspects
from the literature critical to understanding the RGBD odometry algorithm
and subtle variations on published approaches we use that, to our
knowledge, have not been previously published elsewhere.

\subsection{Robust Correspondence Computation in RGBD Image Pairs}

Rather than matching all measured surface points and solving for the
point cloud alignment using ICP \cite{121791} or GICP \cite{Segal-RSS-09},
our odometry reduces the measured data in both frames to a set of
visually-distinct surface locations automatically selected via a user-specified
OpenCV feature detection algorithm, e.g., SIFT \cite{Lowe:2004:DIF:993451.996342},
SURF \cite{Bay:2008:SRF:1370312.1370556} , BRIEF \cite{Calonder:2010:BBR:1888089.1888148},
ORB \cite{Rublee:2011:OEA:2355573.2356268} etc. A user-specified
OpenCV feature descriptor is then computed at each detection location
to create a set of feature descriptors for each RGB image. Robust
methods then compute a correspondence between the RGBD images in a
two-stage process:
\begin{itemize}
\item Stage 1 identifies salient visual matches between the RGB feature
descriptors using a symmetric version of Lowe's ratio test. 
\item Stage 2 refines correspondences from Stage 1 by applying the Random
Sample Consensus (RANSAC) algorithm \cite{Fischler:1981:RSC:358669.358692}
to identify the triplet of corresponding $(X,Y,Z)$ measurements that,
when geometrically aligned, result in the largest amount of inliers,
i.e., geometrically close, pairs in the remaining point set.
\end{itemize}
The two stages operate on distinct subsets measurements from the data.
Stage 1 exploits salient measurements from the RGB image data. Stage
2 exploits salient measurements in the 3D data. While the order of
the stages 1 and 2 could be switched, the computational cost of the
RANSAC algorithm increases significantly when no initial correspondence
is available (see \cite{schnabel-2007-efficient} for RANSAC computational
complexity analysis). This motivates use of RANSAC downstream from
RGB feature matching which, for most OpenCV features, is significantly
faster to compute.

\subsubsection{Stage 1: Robust Image Feature Correspondence via Lowe's Ratio Test
and Symmetry Criteria}

As with many time-based imaging and computer vision algorithms, e.g.,
optical flow, face tracking, and stereo reconstruction from images,
our RGBD odometry approach seeks to find a collection of corresponding
pixels from two sequential images in time. As in other approaches,
we compute sets of visual features from each image, $(F,G)$, that
are assumed invariant to small changes in viewpoint. Correspondences
are found by pairing elements from the feature sets, $(f_{i},g_{j})$.
These correspondences are found by imposing a metric on the feature
space, e.g., a Euclidean metric $\left\Vert f_{i}-g_{j}\right\Vert $,
and then associating features using a criterion on this metric, e.g.,
nearest neighbors: $j=\min_{k}\left(\left\Vert f_{i}-g_{k}\right\Vert ,\,\,g_{k}\in G\right)$.
Unfortunately, application of this type of matching often results
in many incorrect correspondences; especially when many features share
similar values.

In this work we adopt a symmetric version of Lowe's ratio test \cite{Lowe:2004:DIF:993451.996342}\textbf{
}to identify and reject potentially incorrect visual feature correspondences.
Conceptually, this is accomplished by examining the uncertainty in
the match. If computation were not a concern, this uncertainty could
be found by perturbing the value of $f_{i}$ and observing the variability
of the corresponding element $g_{j}$ but this is not feasible in
real-time applications. Lowe's test simplifies this decision by dividing
the distances between $f_{i}$ and it's 2-nearest neighbors $(g_{j},g_{k})$
where $\left\Vert f_{i}-g_{j}\right\Vert \leq\left\Vert f_{i}-g_{k}\right\Vert $
creating the ratio $\frac{\left\Vert f_{i}-g_{j}\right\Vert }{\left\Vert f_{i}-g_{k}\right\Vert }$.
A user-specified salience parameter, $\lambda_{ratio}$, is applied
to reject correspondences that satisfy the inequality $\frac{\left\Vert f_{i}-g_{j}\right\Vert }{\left\Vert f_{i}-g_{k}\right\Vert }>\lambda_{ratio}$.
Features that pass this test ensure the match for $f_{i}$ into the
set of $G$ is salient since the two best candidate matches in $G$
have significantly different distances for their best and next-best
matches. Unfortunately, Lowe's ratio \emph{is asymmetric as it does
not ensure that the match between} $g_{j}$ \emph{into the set} $F$\emph{
is also salient}. Symmetry is achieved by applying the ratio test
first for the element $f_{i}$; ensuring distinct matches from $f_{i}\rightarrow G$
and then again for $g_{j}$; ensuring distinct matches from $g_{j}\rightarrow F$.
Feature matches that satisfy this symmetric version of Lowe's ratio
test are considered to be candidate correspondences between two images.
Results shown in this article were generated by applying this test
with $\lambda_{ratio}=0.8$ which ensures that, for each correspondence,
the 2nd best match, $\left\Vert f_{i}-g_{k}\right\Vert $, has a distance
that is 25\% larger than the best match$\left\Vert f_{i}-g_{j}\right\Vert $.

\subsubsection{Stage 2: Robust 3D $(X,Y,Z)$ Odometry via RANSAC and RANSAC Refinement}

A second stage uses the image locations and depths of the visual correspondences
from Stage 1 to generate a pair of $(X,Y,Z)$ point clouds. Since
the correspondences between points in these clouds are given via the
visual correspondence, it is possible to directly estimate the transformation
in terms of a rotation matrix, $\mathbf{R}$, and translation, $\mathbf{t}$,
that best-aligns the two pointsets in terms of the sum of their squared
distances using solutions to the ``absolute orientation'' problem,
e.g., \cite{Umeyama:1991:LET:105514.105525,Horn87closed-formsolution}.
However, incorrect correspondences may still exist in the data and,
if so, they can introduce significant error. 

The RANSAC algorithm leverages the geometric structure of the measured
3D data to reject these incorrect correspondences. It accomplishes
this by randomly selecting triplets of corresponding points, computing
the transformation,$\left(\mathbf{R},\mathbf{t}\right)$, that best
aligns the triangles these points describe and then scoring the triplet
according to the number of observed inliers in the aligned point clouds.
Inliers are determined by thresholding the geometric distance between
corresponding points after alignment with a user-specified parameter,
$\lambda_{inlier}$, such that a point pair $(\mathbf{p}_{i},\mathbf{p}_{j})$
is marked as an inlier if $\left\Vert \mathbf{p}_{i}-\left(\mathbf{R}\mathbf{p}_{j}+\mathbf{t}\right)\right\Vert <\lambda_{inlier}$.
In this way, the RANSAC-estimated transformation, $\widehat{\left(\mathbf{R},\mathbf{t}\right)}$,
is given by aligning the triplet of points having best geometric agreement
(in terms of inlier count) between the two point clouds. 

We use the RANSAC algorithm included as part of the Point Cloud Library
(PCL v1.7) which also has an option to refine the inlier threshold.
The refinement process allows the RANSAC algorithm to dynamically
reduce the inlier threshold by replacing it with the observed standard
deviation of the inlier distances at the end of each trial. Specifically
the inlier threshold is replaced when the observed variation of the
inliers is smaller than the existing distance threshold, e.g., for
each refinement iteration the new threshold $\hat{\lambda}_{inlier}=\min\left(3\sigma_{inliers},\lambda_{inlier}\right)$.
Results shown in this article are generated using PCL's RANSAC algorithm
(with refinement) with user-specified settings that allow a maximum
of 200 RANSAC iterations and an inlier distance threshold of $\lambda_{inlier}=5cm.$

\subsection{RGBD Cameras: Calibration, 3D Reconstruction, and Noise}

\subsubsection{Calibration}

Primesense sensors are shipped with factory-set calibration parameters
written into their firmware. These parameters characterize the physical
characteristics of the RGB and IR/Depth cameras that make up the RGBD
sensor and allows data from these cameras to be fused to generate
color-attributed $(X,Y,Z)$ point clouds. While it is possible to
use the factory-provided calibration settings, better odometry estimates
are possible using experimentally calibrated parameters. Calibration
describes the process of estimating the \emph{in-situ} image formation
parameters for both the RGB and IR imaging sensors and the transformation,
i.e., rotation and translation, between these two sensors. Calibration
of the intrinsic, i.e., image formation, parameters for each camera,
provide values for the $(x,y)$ focal distance,$(f_{x},f_{y})$, the
image center/principal point, $(c_{x},c_{y}),$ and the $(x,y)$ pixel
position shift, $(\delta_{x},\delta_{y})$, that occur due to radial
and tangential lens distortions. Estimates for these values are obtained
by detecting (via image processing) the $(x,y)$ image positions of
features in images of a calibration pattern having a-priori known
structure (typically a chess board). After intrinsic calibration,
the extrinsic parameters of the sensor pair, i.e, the relative position
and orientation of the RGB and IR sensor, is estimated by applying
a similar procedure to images from both cameras when viewing the same
calibration pattern.

The key benefit of accurate calibration is to reduce the uncertainty
in the location of projected depth and color measurements (intrinsic
parameters) and to align/register the depth and color measurements
from these two sensors to a common $(x,y)$ image coordinate system
(extrinsic parameters) which, for simplicity, is taken to coincide
with the coordinate system of the RGB camera sensor. When using the
factory-shipped calibration values, this registration can occur on-board
the Primesense sensor (hardware registration). Experimentally estimated
calibration parameters are typically more accurate than factory settings
at the expense of additional computational cost (software registration). 

Work in this article uses the factory-provided extrinsic calibration
values for hardware registration of the RGB/IR images and experimentally
calibrated values for the intrinsic/image formation parameters. This
compromise uses RGBD sensor hardware to perform the computationally
costly registration step and software to reconstruct 3D position values.
While use of the factory-provided extrinsic parameters can potentially
sacrifice accuracy, we found these values resulted in similar odometry
and simultaneously significantly reduced the software driver execution
overhead. As RGBD cameras are fixed focal length, the camera calibration
parameters are fixed in time and are recalled by the odometry algorithm
during initialization.

\subsubsection{Point Cloud Reconstruction}

Measured 3D $(X,Y,Z)$ positions of sensed surfaces can be directly
computed from the intrinsic RGBD camera parameters and the measured
depth image values. The $Z$ coordinate is directly taken as the depth
value and the $(X,Y)$ coordinates are computed using the pinhole
camera model. In a typical pinhole camera model 3D $(X,Y,Z)$ points
are projected to $(x,y)$ image locations, e.g., for the image columns
the $x$ image coordinate is $x=f_{x}\frac{X}{Z}+c_{x}-\delta_{x}$.
However, for a depth image, this equation is re-organized to ``back-project''
the depth into the 3D scene and recover the 3D $(X,Y)$ coordinates
as shown by equation (\ref{eq:XYZ_pinhole_reconstruction-1}) 

\begin{center}
\begin{equation}
\begin{array}{ccc}
X & = & (x+\delta_{x}-c_{x})Z/f_{x}\\
Y & = & (y+\delta_{y}-c_{y})Z/f_{y}\\
Z & = & Z
\end{array}\label{eq:XYZ_pinhole_reconstruction-1}
\end{equation}
\par\end{center}

where $Z$ denotes the sensed depth at image position $(x,y)$ and
the remaining values are the intrinsic calibration parameters for
the RGB camera.

\subsubsection*{Measurement Noise}

Studies of accuracy for the Kinect sensor show that assuming a Gaussian
noise model for the normalized disparity provides good fits to observed
measurement errors on planar targets where the distribution parameters
are mean $0$ and standard deviation $\sigma_{d'}=0.5$ pixels \cite{s120201437}.
Since depth is a linear function of disparity, this directly implies
a Gaussian noise model having mean $0$ and standard deviation $\sigma_{Z}=\frac{m}{f_{x}b}Z^{2}\sigma_{d'}$
for depth measurements where $\frac{m}{f_{x}b}=-2.85e^{-3}$ is the
linearized slope for the normalized disparity empirically found in
\cite{s120201437}. Since 3D $(X,Y)$ coordinates are also derived
from the pixel location and the depth, their distributions are also
known as shown below:

\begin{equation}
\begin{array}{ccc}
\sigma_{X} & = & \frac{x_{im}-c_{x}+\delta_{x}}{f_{x}}\sigma_{Z}=\frac{x_{im}-c_{x}+\delta_{x}}{f_{x}}(1.425e^{-3})Z^{2}\\
\sigma_{Y} & = & \frac{y_{im}-c_{y}+\delta_{y}}{f_{y}}\sigma_{Z}=\frac{y_{im}-c_{y}+\delta_{y}}{f_{y}}(1.425e^{-3})Z^{2}\\
\sigma_{Z} & = & \frac{m}{f_{x}b}Z^{2}\sigma_{d'}=(1.425e^{-3})Z^{2}
\end{array}\label{eq:noise_models}
\end{equation}

These equations indicate that 3D coordinate measurement uncertainty
increases as a quadratic function of the depth for all 3 coordinate
values. However, the quadratic coefficient for the $(X,Y)$ coordinate
standard deviation is at most half that in the depth direction, i.e.,
$(\sigma_{X},\sigma_{Y})\approx0.5\sigma_{Z}$ at the image periphery
where $\frac{x-c_{x}}{f}\approx0.5$, and this value is significantly
smaller for pixels close to the optical axis. 

For example, consider a ``standard'' Primesense sensor having no
lens distortion and typical factory-set sensor values: $(f_{x},f_{y})=f=586\pm30$
for focal length, $(640,480)$ for image $(x,y)$ dimension, and $(c_{x},c_{y})=(320,240)$
for the image center. In this case the ratios $(\frac{x-c_{x}+\delta_{x}}{f_{x}},\frac{y-c_{y}+\delta_{y}}{f_{y}})=(0,0)$
at the image center and $(\frac{x-c_{x}+\delta_{x}}{f_{x}},\frac{y-c_{y}+\delta_{y}}{f_{y}})=(0.548\pm0.028,.411\pm0.021)$
at the $(x,y)$ positions on the image boundary. With this in mind,
the $(X,Y,Z)$ coordinates of a depth image are modeled as measurements
from a non-stationary Gaussian process whose mean is $0$ at all points
but whose variance changes based on the value of the triplet $(x,y,Z)$.

\subsection{Odometry Covariance Estimation}

The RGBD reconstruction equations (\ref{eq:XYZ_pinhole_reconstruction-1})
and measurement noise equations (\ref{eq:noise_models}) provide a
statistical model for $(X,Y,Z)$ measurement errors one can expect
to observe in RGBD data. Our approach for odometry parameter covariance
estimation creates two point clouds from the set of inliers identified
by the previously mentioned RANSAC algorithm. We then simulate an
alternate instance of the measured point cloud pair by perturbing
the $(X,Y,Z)$ position of each point with random samples taken from
a $0$ mean Gaussian distribution whose standard deviation is given
by measurement noise equations (\ref{eq:noise_models}). Each pair
of perturbed point clouds is aligned via the solution to the absolute
orientation problem (\cite{Umeyama:1991:LET:105514.105525}) to generate
an estimate of the odometry parameters. For a set of $N$ perturbed
point cloud pairs, let $\xi_{n}$ denote the odometry parameters obtained
by aligning the $n^{th}$ simulated point cloud pair. Unbiased estimates
for the statistical mean, $\widehat{\mu}$, and covariance, $\widehat{\Sigma}$,
of the odometry parameters is computed via:

\[
\widehat{\mu}=\frac{1}{N}\sum_{n=1}^{N}\xi_{n},\,\,\,\,\widehat{\sum}=\frac{1}{N-1}\sum_{n=1}^{N}(\xi_{n}-\widehat{\mu})(\xi_{n}-\widehat{\mu})^{t}
\]

Covariance estimates published in this article are generated by perturbing
each point cloud pair 100 times, i.e., $N=100$.

\section{\label{sec:Methodology}Methodology}

Our approach for odometry seeks to compute instantaneous odometry
directly from sensor measurements in a manner similar to a sensor
driver. As such, there is no attempt for higher-level processing such
as temporal consistency \cite{De-Maeztu:2015:TCG:2849459.2849478}
or keyframing \cite{Leutenegger:2015:KVO:2744155.2744163}. Given
this design goal, such tasks are more appropriate for user-space or
client-level application development where the problem domain will
drive the algorithm choice and design.

The following steps outline of the RGBD odometry driver:
\begin{enumerate}
\item Reduce the dimension of the data by converting the RGB image into
a sparse collection of feature descriptors that characterize salient
pixels in the RGB image.

\begin{enumerate}
\item Discard measurements at the image periphery and measurements having
invalid or out-of-range depth values.
\item Detect feature locations in the RGB image and extract image descriptors
for each feature location.
\end{enumerate}
\item For sequential pairs of RGBD images in time, match their feature descriptors
to compute candidate pixel correspondences between the image pair
using Lowe's Ratio test as discussed in \S~\ref{sec:Background-Information}.
\item Using only the candidate pixel correspondences from (2), compute two
sparse $(X,Y,Z)$ point clouds from the RGBD feature correspondences
using equation (\ref{eq:XYZ_pinhole_reconstruction-1}) from \S~\ref{sec:Background-Information}.
\item Apply Random Sample Consensus (RANSAC) to find the triplet of corresponding
$(X,Y,Z)$ points that, when aligned using the rotation and translation
$\widehat{\left(\mathbf{R},\mathbf{t}\right)}$, maximize the geometric
agreement (number of inliers) of the corresponding point cloud data
as described in \S~\ref{sec:Background-Information}.
\item The subset of points marked as inliers from (4) are used to generate
a new pair of point clouds. We then perturb the $(X,Y,Z)$ data in
each point cloud according to the measurement noise models of equation
(\ref{eq:noise_models}) and empirically compute the 6x6 covariance
matrix, $\widehat{\sum}$, of the odometry parameters as described
in \S~\ref{sec:Background-Information}.
\end{enumerate}
The 6DoF odometry estimate, $\xi$, is taken taken directly from the
alignment transformation parameters, $\widehat{\left(\mathbf{R},\mathbf{t}\right)}$,
computed in step 4. These values characterize the vehicle position
and angular velocities during the time interval spanned by each measured
image pair. Step 5 provides, $\widehat{\Sigma}$, our estimate for
the covariance of the odometry parameters.

\section{\label{sec:Results}Results}

To evaluate our approach for RGBD odometry we implemented the algorithm
as a C++ node using the Robot Operating System (ROS) development framework
\cite{288}. Experiments were conducted using an XTion Pro Live RGBD
camera at full-frame (640x480) resolution and framerate (30Hz). The
frame-to-frame odometry performance is tracked by calibrating the
pose of the RGBD camera to a motion capture system (Optitrack). In
our experiments, we initialize the RGBD camera pose to coincide with
the measured pose as given by averaging 5 seconds (500 samples) of
motion capture data while the RGBD camera is stationary. After initialization,
the pose of the RGBD camera is measured by the motion capture systems
independent from the pose obtained via the time integration of the
frame-to-frame odometry estimates.
\begin{table}
\begin{centering}
\begin{tabular}{|c|c|}
\hline 
Name & Value\tabularnewline
\hline 
\hline 
OpenCV Detector Algorithm & ``ORB''\tabularnewline
\hline 
OpenCV Descriptor Algorithm & ``ORB''\tabularnewline
\hline 
$\lambda_{ratio}$ & 0.8\tabularnewline
\hline 
$\lambda_{inlier}$ & 5cm.\tabularnewline
\hline 
RANSAC MAX ITERATIONS & 200\tabularnewline
\hline 
NUMBER PERTURBATIONS & 100\tabularnewline
\hline 
\end{tabular}
\par\end{centering}
\caption{\label{tab:Algorithm-parameters}Parameters used for the proposed
RGBD odometry algorithm to generate results in this article.}
\end{table}

Using the algorithm parameter values shown in Table \ref{tab:Algorithm-parameters},
our RGBD algorithm runs in real-time (30Hz) on full-frame RGBD data
on a quad-core Intel i5 CPU with a clock speed 2.67GHz and our results
are generated using this configuration and hardware. We have also
run this algorithm, without modification, on an Odroid-XU3 which is
a very small (9.4cm.x7cm.x1.8cm.) and lightweight (\textasciitilde{}
72gram) single board computer having a quad-core Arm7 CPU with a clock
speed of 1.4GHz. Our RGBD algorithm runs on full-frame RGBD data at
a rate of 7.25Hz on this platform and has similar odometry performance.

\begin{figure*}
\begin{centering}
\subfloat[]{\begin{centering}
\includegraphics[width=0.97\textwidth,height=0.8in]{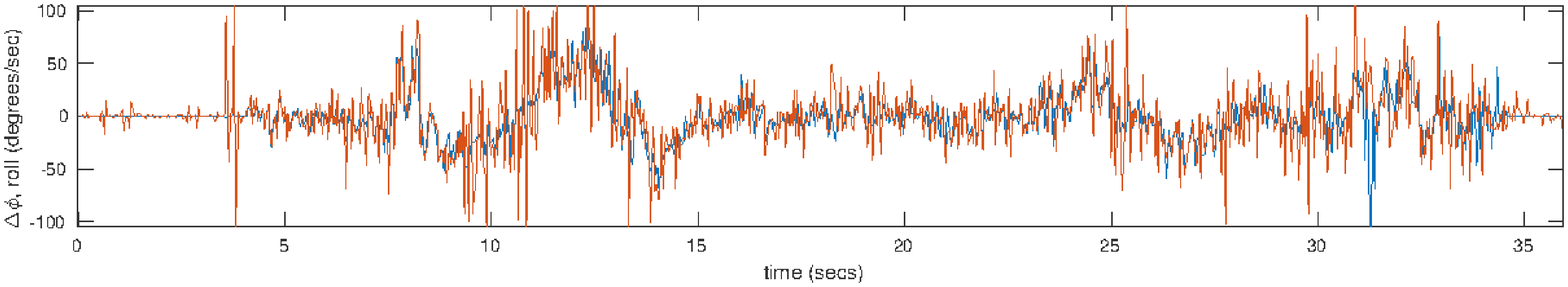}
\par\end{centering}
}
\par\end{centering}
\begin{centering}
\subfloat[]{\begin{centering}
\includegraphics[width=0.96\textwidth,height=0.8in]{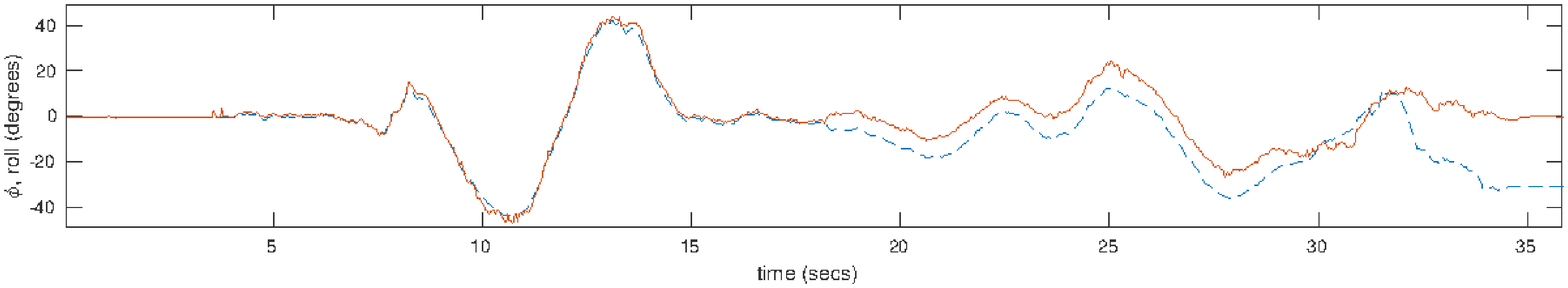}
\par\end{centering}
}
\par\end{centering}
\begin{centering}
\subfloat[]{\begin{centering}
\includegraphics[width=0.95\textwidth,height=0.8in]{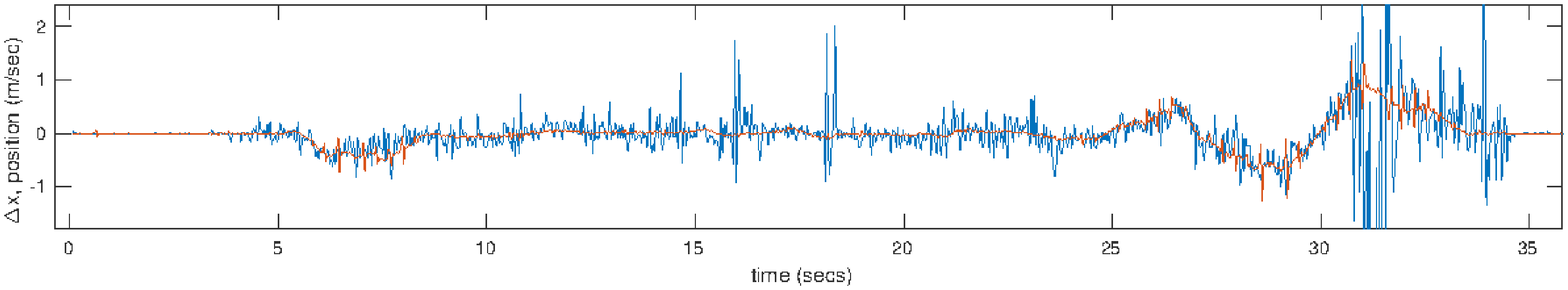}
\par\end{centering}
}
\par\end{centering}
\begin{centering}
\subfloat[]{\begin{centering}
\includegraphics[width=0.97\textwidth,height=0.8in]{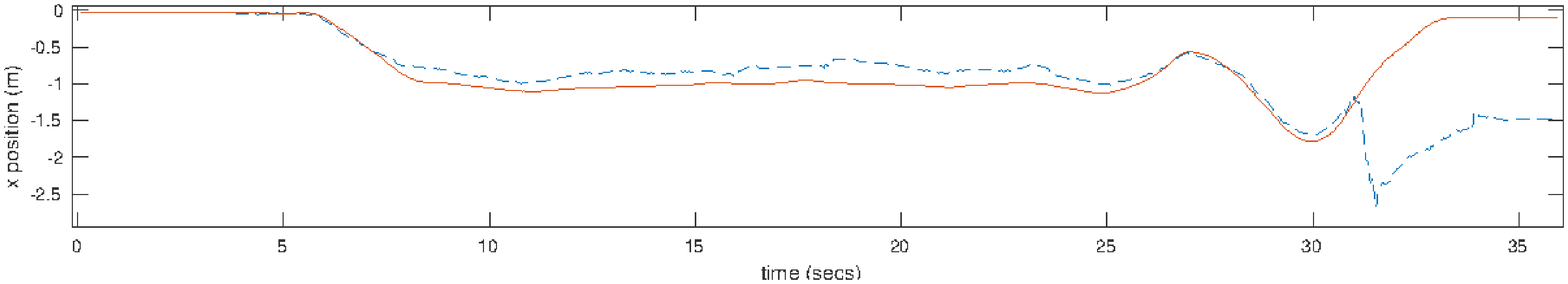}
\par\end{centering}
}
\par\end{centering}
\caption{\label{fig:Estimated-odometry-and-pose}(a,b,c,d) show estimated (blue/dashed)
and motion-capture (orange/solid) values for the velocity and position
of the body-frame $x$-axis during the experiment. (a,b) show the
$x$-axis angular velocity and angular position (roll) respectively.
(c,d) show x position velocities and the absolute x position respectively.
As one would expect, pairs (a,b) and (c,d) have (integral,derivative)
relationships. Inaccurate velocities from visual odometry, e.g., those
at time 32 sec., are coupled with large covariance (see Fig. \ref{fig:Summary}
time index 32 sec.) which indicates to fused-state navigation systems
that these estimates are not reliable.}
\end{figure*}
\begin{figure*}
\begin{centering}
\subfloat[]{\begin{centering}
\includegraphics[width=0.95\textwidth,height=1.2in]{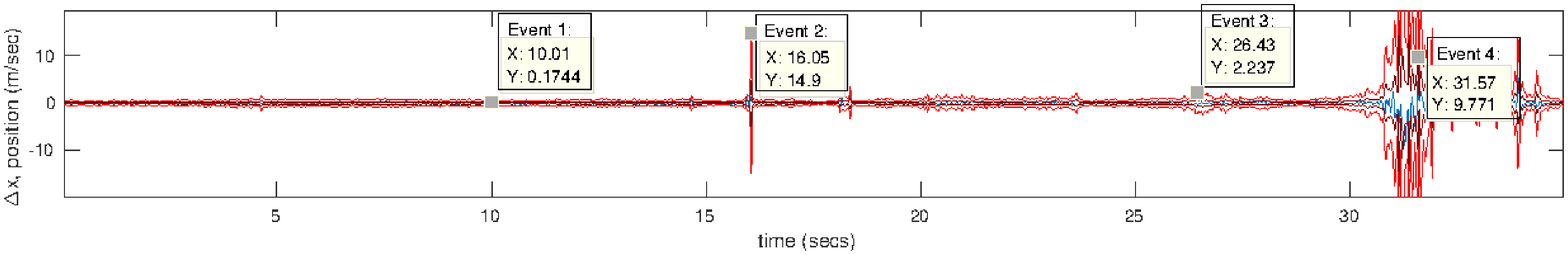}
\par\end{centering}
}
\par\end{centering}
\begin{centering}
\subfloat[]{\begin{centering}
\includegraphics[width=0.48\textwidth,height=0.8in]{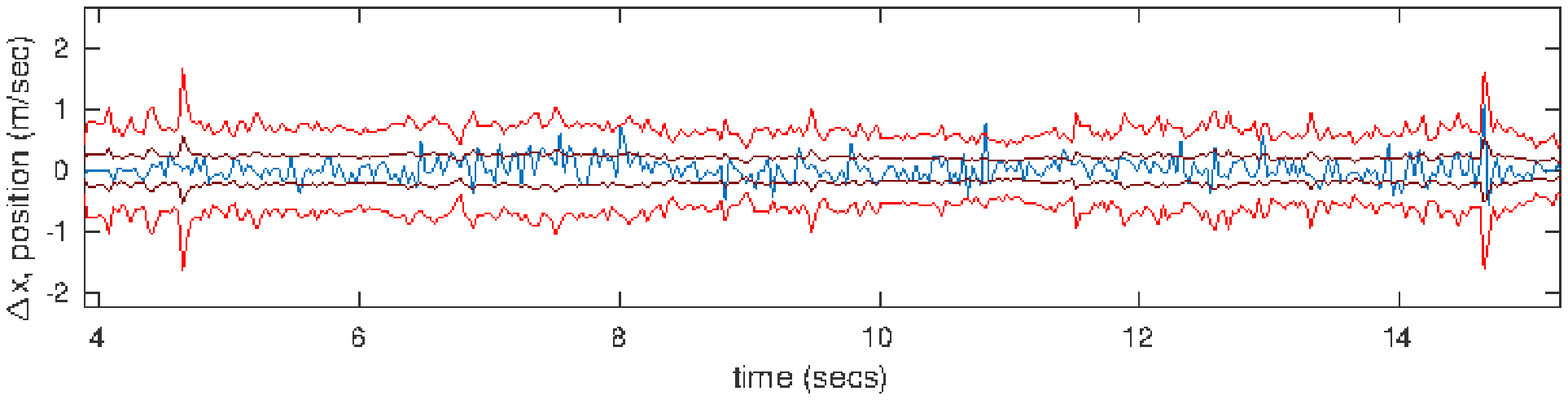}
\par\end{centering}
}\subfloat[]{\begin{centering}
\includegraphics[width=0.48\textwidth,height=0.8in]{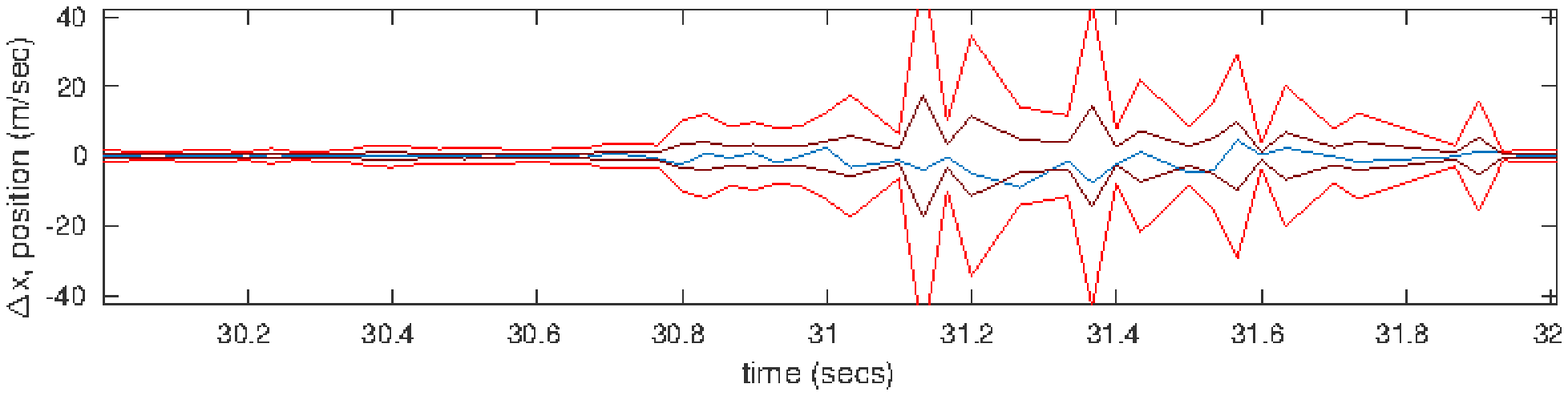}
\par\end{centering}
}
\par\end{centering}
\caption{\label{fig:Summary}Plots (a-c) show three views of odometry error
(blue) and the $\pm1\sigma$ (purple) and $\pm3\sigma$ (red) confidence
intervals for the $x$ position velocity during the 35 second odometry
experiment. (b,c) show close-up views of time intervals (4,14) sec.,
(b), and, (30,32) sec. showing instances having normal odometry, (b),
and highly uncertain odometry, (c). Four time events of interest are
marked on plot (a). Events 1 and 3 mark instances where odometry is
tracking well and parameter variance is smoothly changing. Events
2 and 4 mark instances where the odometry is perceived to be highly-uncertain
and have large associated covariance estimates. As evident in these
plots, covariance values dramatically increase for inaccurate odometry
estimates. This allows fused-state navigation to ignore unreliable
odometry estimates when they occur.}
\end{figure*}

Our results analyze a single 35 second experiment with motion-tracked
RGBD odometry to assess algorithm performance. In this experiment
the sensor views an indoor environment (the laboratory) which includes
office furniture, boxes, chairs and a camera calibration pattern.
The full 6DoF range of motions were generated in the experiment by
manually carrying the camera around the laboratory and includes motions
having simultaneous $(X,Y,Z)$ and (roll, pitch, yaw) variations at
positional velocities approaching 1m/s and angular velocities approaching
50-100 degrees/s. The experiment also includes medium (\textasciitilde{}2m.
range) and long (\textasciitilde{}5m. range) depth images and images
of highly distinct surfaces, e.g., shipping boxes and textured floor
tiles as well as visually confusing surfaces, e.g., a calibration
pattern consisting of regularly spaced black squares on a white background.
All of these contexts promise to provide opportunities to observe
and characterize the odometry parameter estimate accuracy and the
accuracy of the associated covariance of these parameters.
\begin{figure*}
\begin{centering}
\subfloat[Event 1: Accurate odometry: depth \textasciitilde{}3m., slow velocity
(no blurring).]{\begin{centering}
\includegraphics[height=1.3in]{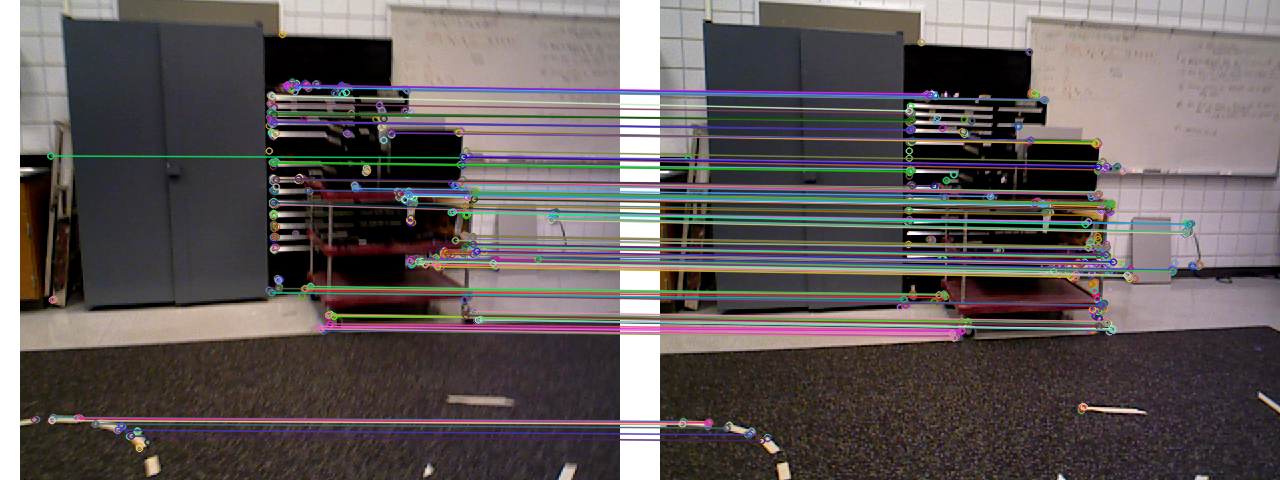}
\par\end{centering}
}\subfloat[Event 2: Odometry error: depth \textasciitilde{}1m., some incorrect
visual correspondences.]{\begin{centering}
\includegraphics[height=1.3in]{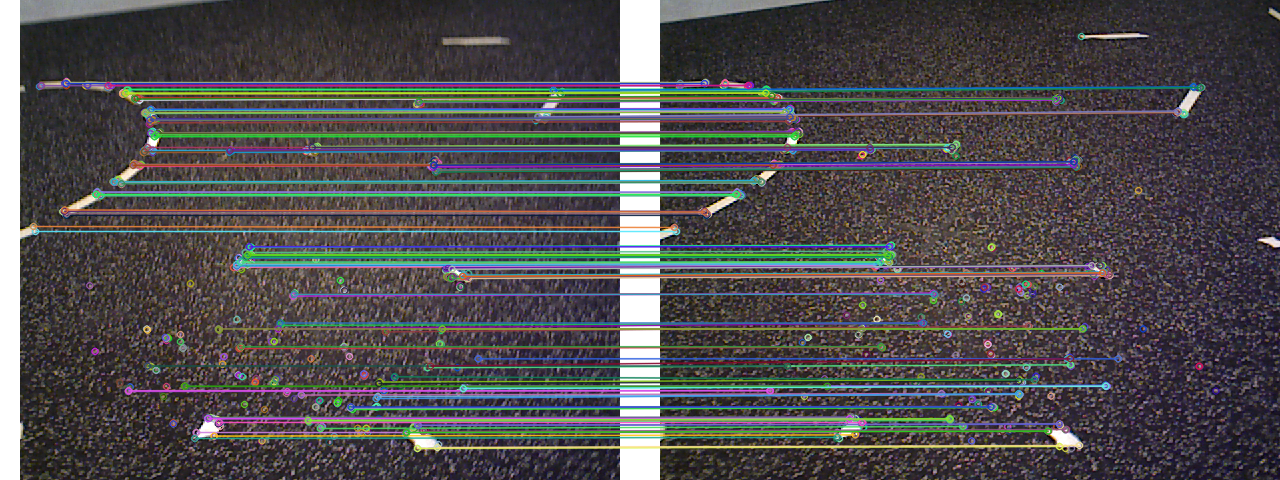}
\par\end{centering}
}
\par\end{centering}
\begin{centering}
\subfloat[Event 3: Accurate odometry: depth \textasciitilde{}3m., higher velocity
(blurring).]{\begin{centering}
\includegraphics[height=1.3in]{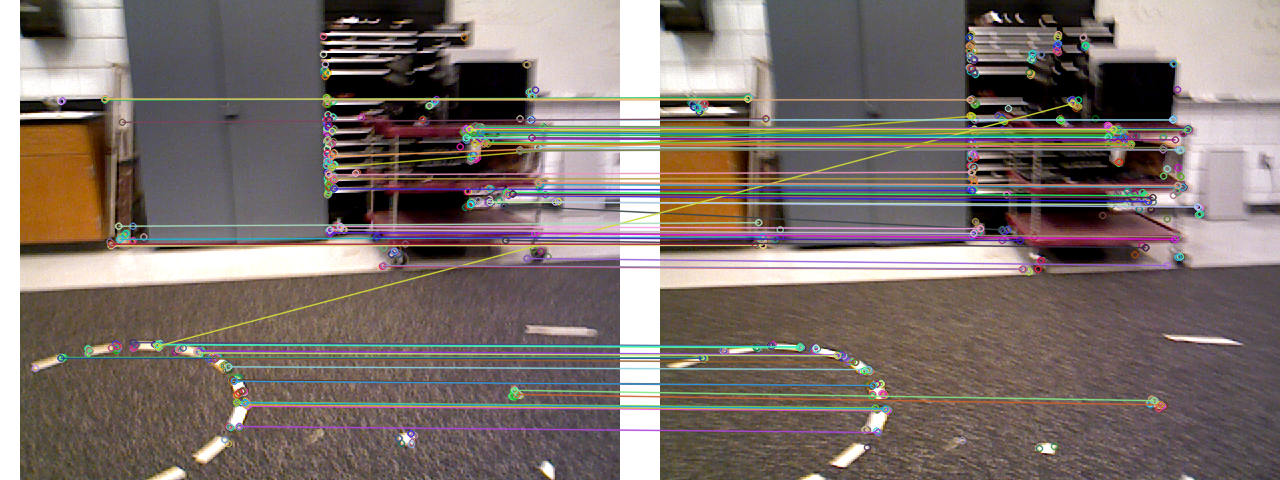}
\par\end{centering}
}\subfloat[Event 4: Worst-case scenario: high velocity, incorrect visual correspondences.]{\begin{centering}
\includegraphics[height=1.3in]{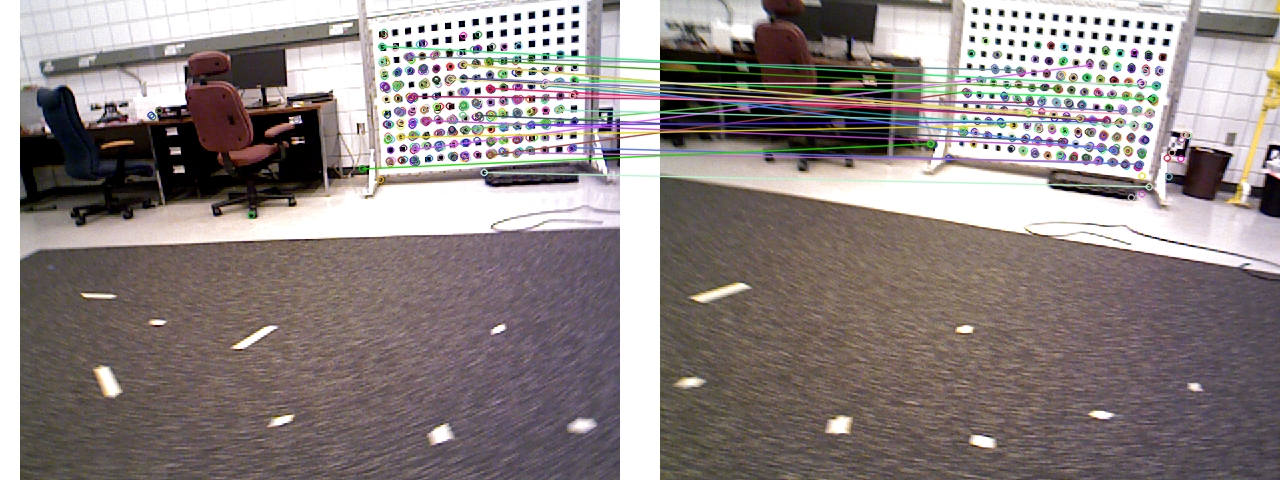}
\par\end{centering}
}
\par\end{centering}
\caption{\label{fig:Odometry-Image-Pairs}(a-d) show RGB image pairs from the
RGBD sensor at Events 1-4 from Fig. \ref{fig:Summary} respectively.
(Left column) Events (1) and (3) show that correct visual correspondences
lead to accurate RGBD odometry. In these cases the estimated odometry
covariance tends to vary smoothly and increases roughly proportional
to vehicle velocity which induces blur to images. (Right column) Events
(2) and (4) show that incorrect visual correspondences lead to inaccurate
RGBD odometry. While this can introduce error to RGBD-only based estimates
for pose (Fig. \ref{fig:Estimated-odometry-and-pose}(b,d)), the estimated
covariance at these time instances increases dramatically (see red
curve of Fig. \ref{fig:Summary}). This allows fused-state navigation
systems to automatically reject these estimates when they occur. As
such, the covariance estimation approach proposed here enables fused-state
navigation systems to benefit from accurate RGBD odometry when it
is available.}
\end{figure*}

Figure \ref{fig:Estimated-odometry-and-pose} depicts estimated (blue/dashed)
and motion-capture (orange/solid) values for the velocity and position
of the of the body-frame x-axis during the experiment. Since motion
capture systems measure absolute pose, i.e., position and orientation,
velocities must be computed by taking the time-differential of the
motion capture data. For this reason, the motion capture angular velocities
are particularly noisy since perturbations in the measured orientation
over short time intervals induce large instantaneous angular velocity
measurements. The close proximity of the estimated and measured velocities
in plots (a,c) show that, in most cases, the RGBD odometry algorithm
tracks well. Errors in angular velocity visible in (a) are localized
to time instants at time indices 16 sec. and 32 sec. The error at
time 16 sec. appears to be due to a combination of a erroneously high
velocity estimate from the motion capture and erroneously low error
from the estimated odometry. The error at time 32 sec. occurs when
the camera makes inaccurate visual matches that cannot be corrected
and result in a failed (highly inaccurate) odometry estimate. A similar
pattern exists for the $x$ position, (d), and position velocities,
(c). As one would expect, odometry errors in (a,c) introduce offsets
between the motion capture and estimated body frame $x$-axis angle
(b) and $x$ position (d) since these quantities are obtained by integrating
the their respective velocity signals (a,c).

Figure \ref{fig:Summary} depicts odometry errors for the vehicle,
i.e., body-frame, $x$-axis (blue) and the estimated $\pm1\sigma$
(purple) and $\pm3\sigma$ (red) confidence intervals for the $x$-axis
velocity parameter. Four events of interest are marked in Fig. \ref{fig:Summary}.
Events (1,3) demonstrate typical, i.e., nominal, RGBD odometry performance
as the vehicle velocity and scene depth simultaneously vary. Events
(2,4) demonstrate atypical, i.e., potentially erroneous, RGBD odometry
and demonstrate that, for both time instances where the odometry is
inaccurate, the covariance dramatically increases.

Analysis of the odometry error indicates that the estimated covariances
underestimate the experimentally measured odometry parameter uncertainty.
This is to be expected since the simulated point clouds from which
this covariance is derived does not account for noise from many sources
such as incorrect visual feature correspondences. Despite this, the
estimated covariances increase and decrease in a manner that is approximately
proportional to the experimentally observed parameter covariance.

Analysis of the data collected shows that taking $\Sigma=9\widehat{\Sigma}$
will place \textasciitilde{}99\% of measurements within their respective
$3\sigma$ bounds for all 6DoF. While the error distributions are
not Gaussian, they behave reasonably well and $\Sigma$ should be
a functional measure for sensor fusion. For this reason, we propose
taking $\Sigma=9\widehat{\Sigma}$ as an estimate for the measurement
uncertainty (Further analysis of the required multiple to place \textasciitilde{}95\%
of sample within $2\sigma$ may be a more practical choice.)

Figure \ref{fig:Summary} shows the associated $\pm3\sigma$ curves
for x position velocity estimates (red). Inspection of the covariance
curve shows smooth variations where the odometry parameters are valid
that tend to increase with vehicle velocity. The curve also includes
large jumps in the covariance, e.g., Events (2,4), where potentially
invalid odometry estimates occur.

Figure \ref{fig:Summary}(a-c) depicts representative image pairs
for Events 1-4 respectively. In each case, the image pair is shown
with 2D visual features indicated by circles on the image. The subset
of visual features that have been matched, as described in Step (2)
of our algorithm (see \S~\ref{sec:Methodology}), have a line shown
that connects corresponding feature locations in the image pair. The
left column shows frame pairs from approximately time index 10 sec.
and 26 sec. At these times the RGBD camera views similar scenes but
has significantly different velocities. The velocity increase is visually
apparent in the blurring artifacts which are absent in Event 1 and
present in Event 3. We feel that image blur due to higher vehicle
velocity contributes to inexact visual feature matches and that the
observed increase in parameter covariance that coincides in time with
higher vehicle velocities is evidence of this phenomena. Events (2,4)
show instances where we have either intentionally (Event 4) or unintentionally
(Event 2) caused the visual odometry estimate to fail. The potential
for failure in visual odometry is a phenomenon shared in differing
degrees by all visual odometry algorithms. In this article, one can
see that the RGBD odometry parameter covariance increases dramatically
in these contexts which allows fused-state navigational systems to
autonomously discard these estimates when they do occur.

\section{\label{sec:Conclusion}Conclusion}

This article proposes a method to estimate 6DoF odometry parameters
and their parameter covariance from RGBD image data in real-time.
The proposed method applies robust methods to accurately identify
pairs of $(X,Y,Z)$ surface measurements in sequential pairs of images.
Using RGBD sensor $(X,Y,Z)$ measurement and measurement noise models,
perturbations are introduced to corresponding pairs of $(X,Y,Z)$
points to simulate plausible alternative odometry estimates for the
same image pair. The covariance of the resulting odometry parameters
serves as an estimate for odometry parameter uncertainty in the form
of a full-rank 6x6 covariance matrix. Observation of the odometry
errors with respect to a calibrated motion capture system indicate
that the estimated covariances underestimate the true uncertainty
of the odometry parameters. This is to be expected since the perturbation
approach applied does not model all sources of uncertainty in the
estimate. While many additional error sources exist, we feel that
modeling uncertainty in correspondences and, to a lesser degree, uncertainty
in the true $(x,y)$ projected pixel position, i.e., $(\delta_{x},\delta_{y})$
from equations (\ref{eq:XYZ_pinhole_reconstruction-1}) show the most
promise for explaining differences observed between the apparent experimental
parameter covariance and our real-time estimate for that value. Despite
this, real-time covariances estimated by the proposed algorithm increase
and decrease in a manner that is approximately proportional to the
experimentally observed parameter covariance. As such, we propose
pre-multiplying estimated covariances by an experimentally motivated
factor ($9\widehat{\Sigma}$). The resulting odometry is real-time,
representative of the true uncertainty and modestly conservative which
makes it ideal for inclusion in typical fused-state odometry estimation
approaches. This dynamic behavior greatly expands the navigational
benefits of RGBD sensors by leveraging good RGBD odometry parameter
estimates when they are available and vice-versa. This is especially
important in situations where a pairwise RGBD odometry fails to return
a plausible motion as evidenced in our experiments which show a dramatic
increases in parameter covariance for these events.

\section{\label{sec:Acknowledgments}Acknowledgment}

This research is sponsored by an AFRL/National Research Council fellowship
and results are made possible by resources made available from AFRL's
Autonomous Vehicles Laboratory at the University of Florida Research
Engineering Education Facility (REEF) in Shalimar, FL.

\bibliographystyle{ieeetr}
\bibliography{2016_ION_PLANS_RealTimeRGBDOdometry}

\begin{thebibliography}{10}

\bibitem{MooreStouchKeneralizedEkf2014}
T.~Moore and D.~Stouch, ``A generalized extended kalman filter implementation
  for the robot operating system,'' in {\em Proceedings of the 13th
  International Conference on Intelligent Autonomous Systems (IAS-13)},
  Springer, July 2014.

\bibitem{6225151}
I.~Dryanovski, C.~Jaramillo, and J.~Xiao, ``Incremental registration of rgb-d
  images,'' in {\em Robotics and Automation (ICRA), 2012 IEEE International
  Conference on}, pp.~1685--1690, May 2012.

\bibitem{121791}
P.~Besl and N.~D. McKay, ``A method for registration of 3-d shapes,'' {\em
  Pattern Analysis and Machine Intelligence, IEEE Transactions on}, vol.~14,
  pp.~239--256, Feb 1992.

\bibitem{Segal-RSS-09}
A.~Segal, D.~Haehnel, and S.~Thrun, ``Generalized-{IC}p,'' in {\em Proceedings
  of Robotics: Science and Systems}, (Seattle, USA), June 2009.

\bibitem{7139931}
Z.~Fang and S.~Scherer, ``Real-time onboard 6dof localization of an indoor mav
  in degraded visual environments using a rgb-d camera,'' in {\em Robotics and
  Automation (ICRA), 2015 IEEE International Conference on}, pp.~5253--5259,
  May 2015.

\bibitem{Thrun:2005:PR:1121596}
S.~Thrun, W.~Burgard, and D.~Fox, {\em Probabilistic Robotics (Intelligent
  Robotics and Autonomous Agents)}.
\newblock The MIT Press, 2005.

\bibitem{6696650}
C.~Kerl, J.~Sturm, and D.~Cremers, ``Dense visual {SLAM} for rgb-d cameras,''
  in {\em Intelligent Robots and Systems (IROS), 2013 IEEE/RSJ International
  Conference on}, pp.~2100--2106, Nov. 2013.

\bibitem{Lowe:2004:DIF:993451.996342}
D.~G. Lowe, ``Distinctive image features from scale-invariant keypoints,'' {\em
  Int. J. Comput. Vision}, vol.~60, pp.~91--110, Nov. 2004.

\bibitem{Bay:2008:SRF:1370312.1370556}
H.~Bay, A.~Ess, T.~Tuytelaars, and L.~Van~Gool, ``Speeded-up robust features
  (surf),'' {\em Comput. Vis. Image Underst.}, vol.~110, pp.~346--359, June
  2008.

\bibitem{Calonder:2010:BBR:1888089.1888148}
M.~Calonder, V.~Lepetit, C.~Strecha, and P.~Fua, ``Brief: Binary robust
  independent elementary features,'' in {\em Proceedings of the 11th European
  Conference on Computer Vision: Part IV}, ECCV'10, (Berlin, Heidelberg),
  pp.~778--792, Springer-Verlag, 2010.

\bibitem{Rublee:2011:OEA:2355573.2356268}
E.~Rublee, V.~Rabaud, K.~Konolige, and G.~Bradski, ``Orb: An efficient
  alternative to sift or surf,'' in {\em Proceedings of the 2011 International
  Conference on Computer Vision}, ICCV '11, (Washington, DC, USA),
  pp.~2564--2571, IEEE Computer Society, 2011.

\bibitem{Fischler:1981:RSC:358669.358692}
M.~A. Fischler and R.~C. Bolles, ``Random sample consensus: A paradigm for
  model fitting with applications to image analysis and automated
  cartography,'' {\em Commun. ACM}, vol.~24, pp.~381--395, June 1981.

\bibitem{schnabel-2007-efficient}
R.~Schnabel, R.~Wahl, and R.~Klein, ``Efficient ransac for point-cloud shape
  detection,'' {\em Computer Graphics Forum}, vol.~26, pp.~214--226, June 2007.

\bibitem{Umeyama:1991:LET:105514.105525}
S.~Umeyama, ``Least-squares estimation of transformation parameters between two
  point patterns,'' {\em IEEE Trans. Pattern Anal. Mach. Intell.}, vol.~13,
  pp.~376--380, Apr. 1991.

\bibitem{Horn87closed-formsolution}
B.~K.~P. Horn, ``Closed-form solution of absolute orientation using unit
  quaternions,'' {\em Journal of the Optical Society of America A}, vol.~4,
  no.~4, pp.~629--642, 1987.

\bibitem{s120201437}
K.~Khoshelham and S.~O. Elberink, ``Accuracy and resolution of kinect depth
  data for indoor mapping applications,'' {\em Sensors}, vol.~12, no.~2,
  p.~1437, 2012.

\bibitem{De-Maeztu:2015:TCG:2849459.2849478}
L.~De-Maeztu, U.~Elordi, M.~Nieto, J.~Barandiaran, and O.~Otaegui, ``A
  temporally consistent grid-based visual odometry framework for multi-core
  architectures,'' {\em J. Real-Time Image Process.}, vol.~10, pp.~759--769,
  Dec. 2015.

\bibitem{Leutenegger:2015:KVO:2744155.2744163}
S.~Leutenegger, S.~Lynen, M.~Bosse, R.~Siegwart, and P.~Furgale,
  ``Keyframe-based visual-inertial odometry using nonlinear optimization,''
  {\em Int. J. Rob. Res.}, vol.~34, pp.~314--334, Mar. 2015.

\bibitem{288}
M.~Quigley, K.~Conley, B.~P. Gerkey, J.~Faust, T.~Foote, J.~Leibs, R.~Wheeler,
  and A.~Y. Ng, ``Ros: an open-source robot operating system,'' in {\em ICRA
  Workshop on Open Source Software}, 2009.

\end{thebibliography}

\end{document}